\PassOptionsToPackage{table,x11names}{xcolor}
\PassOptionsToPackage{capitalize}{cleveref}   

\documentclass[]{fairmeta}

\definecolor{linkColor}{rgb}{0.18,0.39,0.62}
\definecolor{lightgray}{gray}{0.9}

\usepackage[utf8]{inputenc}

\usepackage{amssymb}        
\usepackage{mathtools}      
\usepackage{amsthm}
\usepackage{bbm}
\usepackage{nicefrac}

\usepackage{array}        
\usepackage{arydshln}
\usepackage{makecell}

\usepackage{graphicx}
\usepackage{svg}
\usepackage{wrapfig}
\usepackage{framed}

\usepackage{algorithm}
\usepackage[noend]{algorithmic}

\usepackage{enumitem}
\usepackage{xspace}
\usepackage{pifont}
\usepackage{fontawesome5}

\hypersetup{
  colorlinks = true,
  linkcolor  = linkColor,
  citecolor  = linkColor,
  filecolor  = linkColor,
  urlcolor   = linkColor,
}

\theoremstyle{plain}

\theoremstyle{definition}

\theoremstyle{remark}

\newif\ifdraft
\drafttrue

\ifdraft

\else

\fi

\newcommand\our{Text-AB}

\makeatletter
\newcommand*{\@rowstyle}{}
\newcommand*{\rowstyle}[1]{
  \gdef\@rowstyle{#1}%
  \@rowstyle\ignorespaces%
}
\newcolumntype{=}{
  >{\gdef\@rowstyle{}}%
}
\newcolumntype{+}{
  >{\@rowstyle}%
}
\makeatother

\title{Alignment-Free Text-Audiobox for Voice Dubbing and Full-Duplex Dialogue Synthesis}

\author[*]{Sanyuan Chen}
\author[*]{Min-Jae Hwang}
\author[*]{Sho Inoue}
\author[]{Anna Sun}
\author[]{Bokai Yu}
\author[]{David Kant}
\author[]{Dongmin Hyun}
\author[]{Dorian Desblancs}
\author[]{Gregory Antonovsky}
\author[]{Oleg Repin}
\author[]{Peng-Jen Chen}
\author[]{Xutai Ma}
\author[]{Zehai Tu}
\author[]{Juan Pino}
\author[\dagger]{Wei-Ning Hsu}

\affiliation[]{FAIR at Meta}

\contribution[*]{Equal contribution}
\contribution[\dagger]{Senior author}

\abstract{
    We present \textit{Alignment-Free Text-Audiobox} (\our), a unified framework for high-quality voice dubbing and full-duplex dialogue synthesis. 
    Building on a Diffusion Transformer backbone trained with a flow-matching objective, \our\ departs from prior Audiobox system along three key dimensions. 
    First, it operates in a latent diffusion framework using DAC-VAE features that encode 48\,kHz waveforms into a 25\,Hz low-rate latent sequence, providing over 10$\times$ higher compression than previous EnCodec representations while improving resynthesis quality. 
    Second, \our\ is \emph{alignment-free}: it consumes raw text via an off-the-shelf text encoder and learns text--speech alignment through cross-attention, removing the need for forced alignment and explicit duration prediction. 
    Third, we scale the model and data substantially, pretraining a 3B-parameter model on 480k hours of monolingual speech, followed by supervised fine-tuning on three downstream tasks---\textit{cross-lingual voice dubbing}, \textit{full-duplex dialogue synthesis}, and \textit{emotional full-duplex dialogue synthesis}.
    At inference time, \our\ supports both one-shot generation for up to $\sim$1\,min speech and arbitrarily long-form generation via a multi-diffusion scheme.
    We further incorporate a \textit{multi-stage reranking strategy} to effectively enhance generation quality based on automated metrics.
    On a real-world dubbing benchmark, \our\ delivers a step-change improvement over the latest internal dubbing system, with large gains in prosody similarity, voice similarity, naturalness, and shareability. 
    For full-duplex dialogue synthesis, Text-AB approaches human recordings on short-form conversations and substantially outperforms the latest internal model on long-form human-likeness and expressivity, while natively modeling turn-taking, back-channeling, and emotional dynamics. 
    For emotional dialogue synthesis, \our\ with emotion conditioning significantly improves emotion alignment and emotional interaction quality over the baseline without emotion conditioning.
}

\date{\today}
\correspondence{\email{mjhwang@meta.com}}

\begin{document}

\maketitle

\section{Introduction}

Text-to-speech (TTS) aims to generate high-quality speech from text with high clarity and intelligibility. Driven by recent advances in deep learning, TTS systems have made remarkable progress in naturalness and robustness~\citep{DBLP:conf/icassp/ShenPWSJYCZWRSA18,DBLP:conf/aaai/Li0LZL19,DBLP:conf/nips/RenRTQZZL19}. More recently, \emph{zero-shot} TTS has attracted increasing attention, where the model is required to synthesize speech for previously unseen speakers given only a short enrollment utterance at inference time. A variety of approaches have been proposed, including language-model-based methods~\citep{chen2025neural}, diffusion-based methods~\citep{ju2024naturalspeech}, and flow-matching-based methods~\citep{le2024voicebox, vyas2023audiobox}, with some systems achieving near human-parity performance under certain conditions~\citep{chen2024vall}.

Despite these advances in monolingual single-speaker TTS, more complex settings such as multilingual generation for voice dubbing and full-duplex dialogue synthesis remain highly challenging. 
Voice dubbing aims to translate speech into a different language while preserving the speaker’s voice, emotion, and expressiveness from a source audio prompt. A typical industrial solution is a cascaded pipeline: an ASR model transcribes the source speech, a machine translation model produces the target-language transcript, and a TTS model finally generates the target speech~\citep{yang2020large,federico2020speech,artioli2025generative}. In this work we focus on the last component. From a modeling perspective, voice dubbing can be viewed as a cross-lingual zero-shot TTS problem: the input speech serves as an audio prompt, and the model aims to generate target-language speech conditioned on a given target transcript while preserving the source speaker’s characteristics. A major challenge is that large-scale training data are typically monolingual, whereas the use case is cross-lingual, so the model must learn to transfer speaker attributes across languages despite this training–inference mismatch.

On the other hand, full-duplex dialogue synthesis, whose goal is to generate a natural conversation between two speakers, has become an important problem with the rise of full-duplex speech language models (FD-SLMs)~\citep{defossez2024moshi,cui2025recent}.
Training FD-SLM purely on real full-duplex recordings (two-channel conversations from real speakers) satisfies stringent audio-quality requirements, but offers limited control over content, personality, style, and identity. 
Moreover, increasing the proportion of such data in the training mix of FD-SLM can also degrade factuality and lead to identity confusion.
To address this, prior work has relied on synthetic pipelines by generating single-turn monologues using a TTS model, then algorithmically stitching them into a dialogue~\citep{guo2021conversational,xue2023m,lee2023dailytalk,hu2024fctalker,liu2024generative,xie2025fireredtts}. 
Although this approach is straightforward, it fundamentally limits the ability to create highly fluid conversations because it lacks joint modeling of conversational context and turn dynamics.
Recent work has attempted to generate full-duplex dialogue in a more integrated manner~\citep{peng2025vibevoice,zhang2025covomix2,zhu2025zipvoice}. However, these approaches either lack zero-shot voice prompting capabilities or are trained on limited data with modest model scales, leaving room for improvement in both controllability and quality of the  synthesized dialogues.

In this work, we present \emph{Alignment-Free Text-Audiobox} (\our), a unified framework with two variants, \emph{Text-AB-Mono} and \emph{Text-AB-Stereo}, for end-to-end single- and two-channel speech synthesis, respectively. \our\ builds on Audiobox~\citep{vyas2023audiobox}, which also uses a Diffusion Transformer (DiT) backbone~\citep{peebles2023scalable} and a flow-matching objective~\citep{lipman2022flow}, but introduces several key improvements.
First, we adopt a latent diffusion framework~\citep{rombach2022high} with DAC-VAE audio features~\citep{polyak2024movie} that encode 48\,kHz waveforms into a low-rate (25\,Hz) latent sequence. Compared to the EnCodec features used in Audiobox, these representations offer much higher compression (1920$\times$ vs.\ 160$\times$--320$\times$), higher audio resolution (48\,kHz vs.\ 16--24\,kHz), and better resynthesis quality. 
Second, \our\ is an alignment-free, end-to-end model that avoids explicit token-duration prediction, which brings three major benefits: (a) prior work typically relies on regression-based duration predictors that tend to underfit and limit expressivity, while autoregressive or diffusion-based duration models are often less stable in practice; (b) training a duration predictor requires a forced aligner to estimate token durations, and forced-alignment errors can significantly hurt performance, especially on noisy or highly conversational speech; and (c) \our\ uses a single generative component, which can be optimized end-to-end and scaled more easily. 
Third, \our\ is substantially larger, uses stronger audio representations, and is trained on more and higher-quality data with improved training strategies. 

Our main model has 3B parameters ($\sim$10$\times$ larger than Audiobox) and uses a multi-stage training pipeline: 
(i) large-scale pretraining on 480k hours of monolingual data ($\sim$3$\times$ the Audiobox scale) to learn general acoustic and prosodic characteristics, 
(ii) fine-tuning on 2k hours of high-quality monolingual and cross-lingual data for voice dubbing, 
(iii) task-specific fine-tuning on 28k hours of two-channel dialogue data for full-duplex dialogue synthesis, or with additional emotion conditioning for emotional dialogue synthesis.
At inference time, we extend the model to long-form generation via multi-diffusion, which enables arbitrarily long speech generation and removes the effective length constraints imposed by the training data duration distribution.
Moreover, we dramatically increase generation performance by incorporating \textit{multi-stage reranking strategy}, which automatically selects the best output among multiple candidates based on the word error rate and speaker similarity scores.


We evaluate \our\ on three downstream tasks---voice dubbing, full-duplex dialogue synthesis, and emotional full-duplex dialogue synthesis. 
For voice dubbing, Text-AB delivers a step-change improvement over the latest internal dubbing model in human evaluations using a $[-3, 3]$ mean opinion score (MOS) scale: we observe large gains not only in overall shareability (+0.39), but also across fine-grained aspects, including prosody similarity (+0.34), voice similarity (+0.32), and voice naturalness (+0.42). 
For full-duplex dialogue synthesis, \our\ can generate up to 1 minute of audio in a single one-shot pass and supports long-form generation beyond 10 minutes via a multi-diffusion scheme. 
\our\ closely matches real recordings on short-form human-written scripts (only $-0.09$ in overall human-likeness on a 5-point MOS scale) and substantially outperforms the previous dialogue synthesis system on long-form synthetic scripts (+0.86 in overall human-likeness). 
\our\ learns turn dynamics, back-channeling, and emotional evolution directly from data, and therefore does not rely on ad-hoc turn stitching, heuristic back-channel insertion, or explicit emotion tagging used in previous systems.
For emotional full-duplex dialogue synthesis, \our\ additionally incorporates explicit turn-level emotion style embeddings together with turn-level transcriptions as conditioning inputs, allowing the generated emotion to differ from the prompt speaking style.

\section{\our}

\subsection{Flow-Matching}
\label{sec:fm_formula}
Following prior work on Audiobox~\citep{vyas2023audiobox}, we train \our\ model using the flow-matching loss \citep{lipman2022flow}.
Flow matching generates samples from a target data distribution by iteratively transforming
samples drawn from a simple prior distribution, e.g., a standard Gaussian.

During training, given an audio sample in the latent space $\mathbf{X}_1$, we sample a time
step $t \in [0, 1]$ from a logit-normal distribution and a noise sample $\mathbf{X}_0 \sim \mathcal{N}(\mathbf{0}, \mathbf{I})$.
Then, we use the linear interpolation or the optimal-transport path~\citep{lipman2022flow}
to construct the noised sample $\mathbf{X}_t = t \mathbf{X}_1 + \bigl(1 - (1 - \sigma_{\min}) t \bigr)\mathbf{X}_0$, where $\sigma_{\min} = 10^{-5}$.
The model is trained to predict the velocity $\mathbf{V}_t = \frac{d\mathbf{X}_t}{dt}= \mathbf{X}_1 - (1 - \sigma_{\min}) \mathbf{X}_0$, which determines how to move $\mathbf{X}_t$ toward the target sample $\mathbf{X}_1$.
Let $\theta$ denote the model parameters and $\mathbf{C}$ the conditions.
We denote the model's predicted velocity by $u(\mathbf{X}_t, \mathbf{C}, t; \theta)$.
The training objective is the mean squared error between the ground-truth and predicted
velocities:
\begin{equation}
\mathcal{L}_{\text{FM}}(\theta)
= \mathbb{E}_{t, \mathbf{X}_0, \mathbf{X}_1, \mathbf{C}}
\left[ \left\| u(\mathbf{X}_t, \mathbf{C}, t; \theta) - \mathbf{V}_t \right\|_2^2 \right].
\end{equation}

At inference time, we first sample $\mathbf{X}_0 \sim \mathcal{N}(\mathbf{0}, \mathbf{I})$
and then solve the corresponding ordinary differential equation (ODE) to obtain
$\mathbf{X}_1$ using the model's estimate of $d\mathbf{X}_t/dt$. We employ a simple
first-order Euler ODE solver with a fixed set of $N$ time steps tailored to our model.

\subsection{Architecture}
\label{sec:arch}

\begin{figure*}[h]
	\centering
	\includegraphics[width=1.0\textwidth]{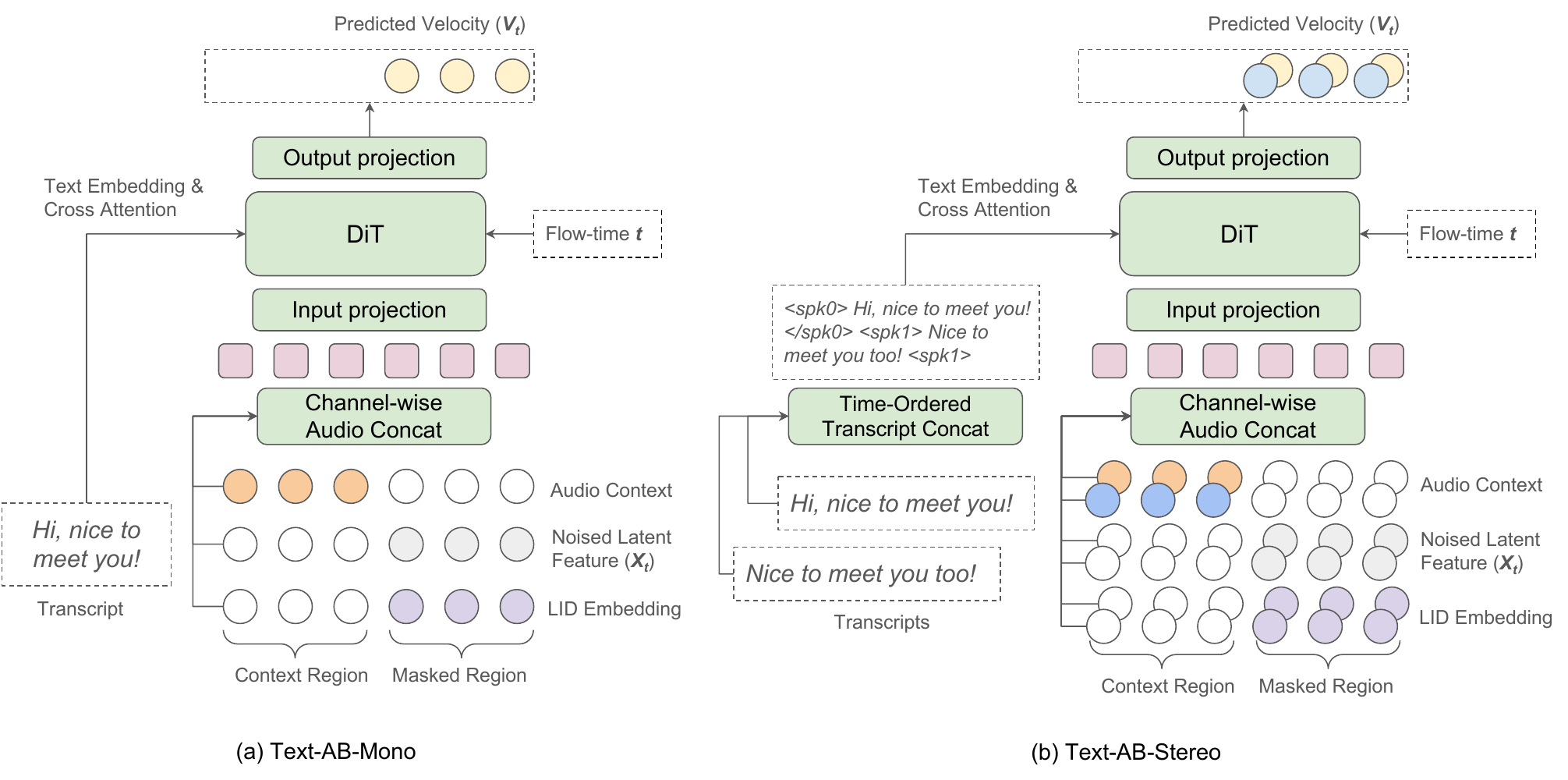}
	\caption{
        The architectures of \our: (a) Text-AB-Mono and (b) Text-AB-Stereo.
        \label{fig:arc}
    }
\end{figure*}%

Figure~\ref{fig:arc} illustrates the architectures of \our\, which includes the two variants---Text-AB-Mono and Text-AB-Stereo for single and two channel speech generation, respectively.

\paragraph{\textbf{DiT Backbone.}}
The model is based on a DiT backbone~\citep{peebles2023scalable}. In each Transformer block~\citep{vaswani2017attention}, the flow-time embedding modulates (i) the outputs of normalization layers via scale and bias, and (ii) the outputs of self-attention and feed-forward layers via scale. A multi-layer perceptron (MLP) takes the flow-time embedding as input and predicts six modulation parameters (four scales and two biases). Similar to \cite{polyak2024movie}, the MLP is shared across all layers and only layer-dependent biases are added to the MLP outputs, which saves parameters without sacrificing performance.

\paragraph{\textbf{Audio Latent.}}
We adopt a latent diffusion framework~\citep{rombach2022high} in which audio is represented as compact latent features. Specifically, we use the DAC-VAE model \citep{polyak2024movie} to produce latent features at a 25~Hz frame rate and 128 dimensions for 48~kHz input audio.
Following Audiobox, we take partially masked noised audio features as input and additionally use an audio context to enable audio prompting.  The audio context is also represented as the DAC-VAE features of the audio and replaced with zero vectors in the masked region. To perform audio generation without any audio context, we simply input an all-zero sequence as the audio context. We concatenate it with the noised audio latent along the channel dimension frame by frame.

\paragraph{\textbf{Text and Language Condition.}}
In contrast to Audiobox, which relies on force-aligned text tokens obtained from a force-aligner, our model directly consumes raw text and implicitly learns text–speech alignment using cross-attention. Concretely, we obtain high-level text embeddings using an mT5 text encoder \citep{xue2021mt5} and feed these embeddings into cross-attention layers within the Transformer encoder.  To support multilingual speech generation, we additionally condition on frame-level language IDs. We first obtain frame-level language labels, then convert them to embedding vectors using a simple embedding layer. These embeddings are concatenated with the audio context and noised audio features along the channel dimension.

\paragraph{\textbf{Context Zero-Out.}}

Note that we optimize the predicted velocity only within the masked region following Audiobox, which leads to a discrepancy between training and inference. 
During training, the context region of the noised audio feature is obtained via interpolation between ground-truth speech and random noise, as detailed in \cref{sec:fm_formula}. However, during inference, it is derived from model predictions generated in previous ODE steps. Since the model is not optimized to predict velocity for the context region, these predictions may be inaccurate, leading to error accumulation in the noised audio features across subsequent ODE steps. To mitigate this mismatch, we mask the context region with zero values during both phases. Notably, this does not hinder performance, as the necessary contextual information is already provided via the auxiliary audio context.


Similarly, the mismatch would also occur in the language-ID embeddings. 
More specifically, in the monolingual training data, the context region of the language-ID embedding always matches the target region.
During inference, however, the language in the context region may differ from the target region (e.g., Spanish in the context and English in the target). 
To avoid this issue, we likewise ignore the context region of the language-ID embeddings in both training and inference by masking them with zero-valued tensors, as illustrated in Figure~\ref{fig:arc}.




\paragraph{\textbf{Stereo Speech Modeling.}}
To support stereo speech generation in Text-AB-Stereo, we concatenate the audio context, noised audio features, and language-ID embeddings from the two channels along the channel dimension, and let the model jointly predict the velocities for both channels. 
For the text input, we construct a single transcript by temporally ordering and concatenating the two speaker transcripts, inserting special tokens to denote the speaker ID of each segment, and feed this unified transcript to the text encoder.



    


\section{Training}
\label{sec:train}

\subsection{Pre-Training}
We first conduct large-scale pretraining on Text-AB-Mono using multilingual monologue data and a speech-infilling task, following Audiobox. More specifically, we randomly mask DAC-VAE features and train the DiT model to predict the velocity of the features distribution in the masked regions using the flow-matching loss. We use 480k hours of speech data consisting of 380k hours of English (En) data and 100k hours of Spanish (Es) data for this pretraining stage.
Starting from this pretrained Text-AB-Mono model, we first perform dubbing SFT (Section~\ref{sec:dubbing_sft}). Then, we branch into dialogue SFT (Section~\ref{sec:dialog_sft}) and emotional dialogue SFT (Section~\ref{sec:emodialog_sft}) for full-duplex dialogue synthesis and emotional dialogue synthesis, respectively.

\subsection{Dubbing SFT}
\label{sec:dubbing_sft}

\paragraph{\textbf{Sentence-level masking.}}
Unlike pretraining, which applies random masking within a single sentence, the inference setting requires sentence-level masking, where source and target sentences are concatenated along the time dimension and only the target sentence is masked. 
To bridge this gap, we design a sentence-level masking strategy. 
Specifically, we constructed a 2k hours of multi-sentence SFT dataset containing multiple complete sentences for both En and Es (1k hours for each language). 
During training, we randomly sample two sets of consecutive sentences from this dataset and use them as the left and right parts of the model input. 
For the loss computation, we apply masks only to the right part, closely mimicking the inference scenario where only the target sentence is masked. 

\paragraph{\textbf{Cross-lingual SFT data synthesis.}}

Dubbing systems structurally require cross-lingual TTS, where the audio prompts and target text are in different languages. 
To reflect this environment during training, we adopt a Unit-VoiceBox~\citep{barrault2023seamless} as a cross-lingual voice conversion task, 
artificially generating a 50 hours of voice and style-aligned cross-lingual SFT data between En and Es.
We mixed the cross-lingual SFT data with multi-sentence SFT data, resulting in a total of 2.1k hours of SFT data.

\subsection{Dialogue SFT}
\label{sec:dialog_sft}
Given the pretrained Text-AB-Mono model, we perform Dialogue SFT on stereo full-duplex dialogue data to obtain the full-duplex dialogue synthesis model. 
To initialize Text-AB-Stereo from Text-AB-Mono, we modify only the input and output projection layers by duplicating both projection matrices to accommodate the two-channel input and output. In particular, the input projection weights are divided by 2 so that the distribution of the projected features remains consistent with pretraining.
We fine-tune on 28k hours of English two-channel full-duplex dialogue data, consisting of real conversational recordings between two speakers captured on separate channels. 
For the speech conditioning, we concatenate the speech features of both speakers along the channel dimension. 
For the text conditioning, in contrast, we concatenate the transcripts of both speakers in temporal order with special speaker-ID tokens delimiting each turn, as described in Section~\ref{sec:arch}. 
This unified transcript allows the model to attend to the full conversational context when generating each frame.

\subsection{Emotional dialogue SFT}
\label{sec:emodialog_sft}
Following the same approach as dialogue SFT, we fine-tune Text-AB-Mono to support explicit turn-level emotion control. We incorporate emotion style embeddings derived from valence-arousal-dominance (VAD) representations~\citep{russell1977evidence}, which capture emotion along three continuous dimensions: valence, arousal, and dominance. For each turn in the training data, we extract a turn-level VAD vector from the corresponding speech segment using a pretrained wav2vec2-based emotion recognizer~\citep{baevski2020wav2vec}\footnote{\label{fn:vad}Valence-Arousal-Dominance extractor: \url{https://huggingface.co/audeering/wav2vec2-large-robust-12-ft-emotion-msp-dim}}, and project it into the model's hidden dimension via a learned linear layer. The resulting emotion embedding is concatenated with the text encoder output along the sequence dimension and fed into the cross-attention layers of the DiT backbone.
Note that this design decouples emotional expression from the audio prompt's speaking style, enabling the model to generate speech whose emotion differs from the reference speaker's tone while preserving the speaker's voice identity.


\section{Inference}
At inference time, Text-AB supports two modes: (i) one-shot inference, which generates up to about one minute of speech in a single chunk (beyond which quality degrades notably in our experiments), and (ii) long-form inference with a multi-diffusion mechanism, which can generate arbitrarily long speech. 
In this section, we use Text-AB-Stereo as the example to illustrate both inference modes without loss of generality, as Text-AB-Mono can be viewed as a special case where the number of channels is reduced from two to one.

\paragraph{\textbf{One-Shot Inference.}}

\begin{figure*}[h]
	\centering
	\includegraphics[width=1.0\textwidth]{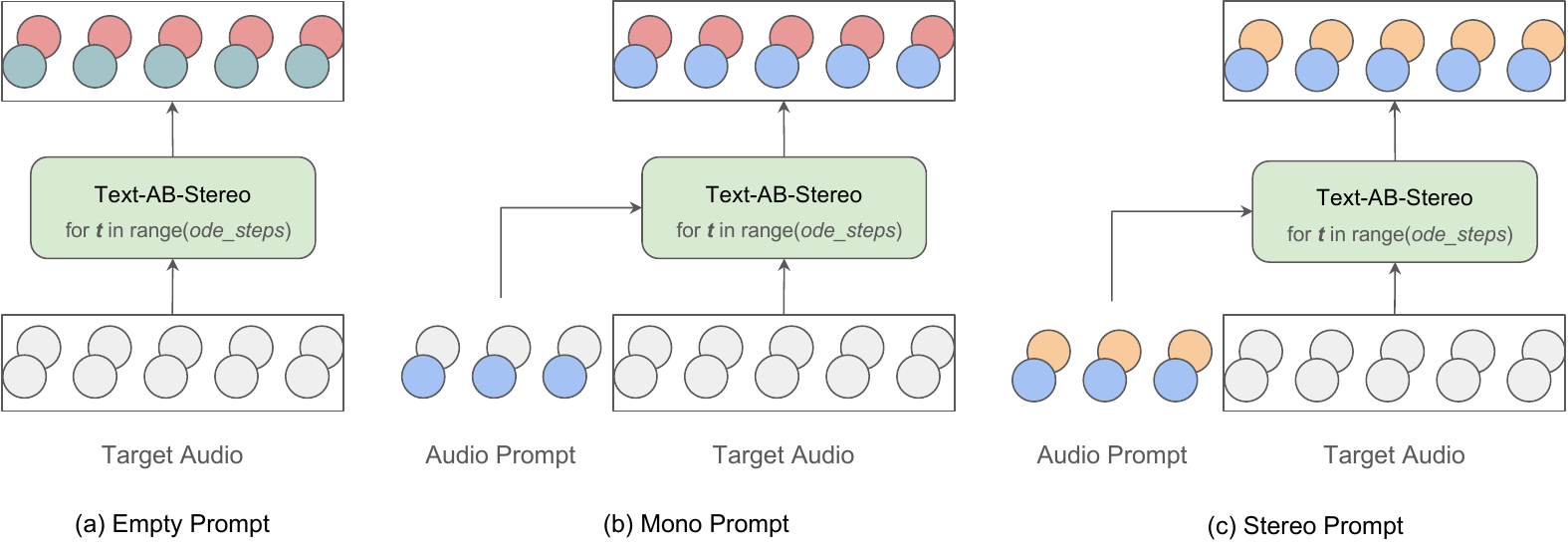}
	\caption{One-Shot inference with different audio prompts.
\label{fig:infer}
    }
\end{figure*}%
Figure~\ref{fig:infer} illustrates one-shot inference under different audio-prompt configurations. In the \textit{empty prompt} setting, no audio prompt is provided. We initialize from random noise and solve the ODE by running the Text-AB-Stereo model forward for a fixed number of ODE steps (denoted by \textit{ode\_steps}), obtaining a two-channel audio with randomly sampled voices on both channels. In the \textit{mono prompt} setting, the input audio prompt contains speech only on one channel. The model then generates speech for both channels, reusing the voice in the prompted channel and automatically creating a random voice for the silent channel. In the \textit{stereo prompt} setting, the input audio prompt contains speech on both channels, corresponding to two speakers, and the model generates a dialogue with two channels that remain consistent with the respective speaker identities.

\paragraph{\textbf{Long-Form Inference.}}

\begin{figure*}[h]
	\centering
	\includegraphics[width=1.0\textwidth]{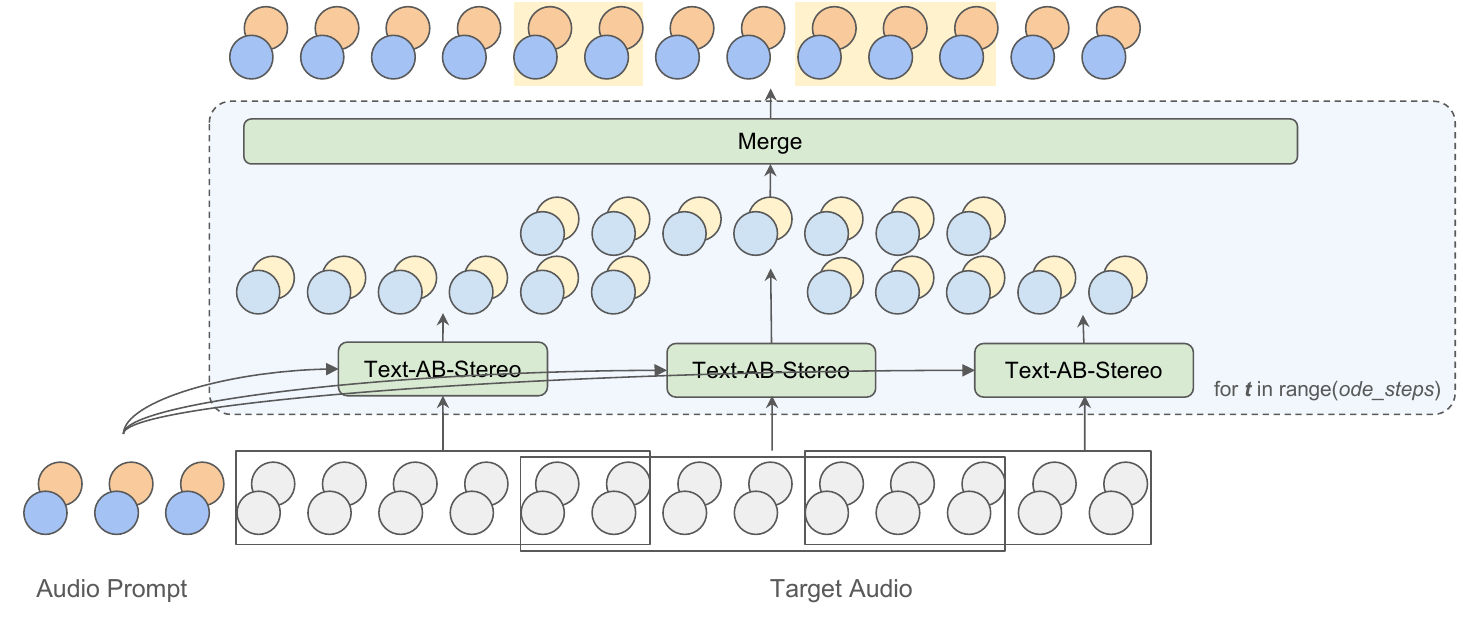}
	\caption{ Illustration of multi-diffusion. A target audio of 13 frames is split into 6, 7, 5 frames, with 2 and 3 frames overlapping. The overlapping frames (highlighted with yellow background at output) are consolidated with weighted average from contributing chunks.
\label{fig:md}
    }
\end{figure*}%

To generate longer conversations, simply concatenating multiple one-minute dialogues produced by Text-AB-Stereo leads to audible artifacts and discontinuities around the chunk boundaries. To address this, we adopt the multi-diffusion inference scheme from Movie Gen Audio~\citep{polyak2024movie} and extend it to variable-length chunking, as illustrated in Figure~\ref{fig:md}. Given a target sequence and an overlapping chunk partition, each chunk is processed independently by Text-AB-Stereo, and at every ODE step the predicted flows in the overlapping frames are merged via a weighted average. This yields a temporally consistent long-form dialogue without artifacts at chunk boundaries. Multi-diffusion is only effective with stereo prompts. When using a mono or empty prompt, we observe that the speaker identity can become inconsistent across chunks for any channel whose speaker is not fixed by the prompt. A simple remedy is to first generate an initial chunk in which both speakers speak, and then use this chunk as the stereo audio prompt for subsequent multi-diffusion inference.

\paragraph{\textbf{Multi-Stage Reranking}}


Motivated by the observation that flow-matching objective enables diverse generations across random seeds, we adopt an objective-metric-based reranking strategy~\citep{chen2024vall, chen2025neural} to automatically select the best sample among multiple candidates for both voice dubbing and full-duplex dialogue synthesis.
More specifically, we first generate multiple audio samples using different random seeds. 
Then, for each sample we compute speaker similarity (SpkSim) between the generated speech and the audio prompt and word error rate (WER) between the ASR transcript of the output speech and the target text, using the WavLM speaker verification model \citep{chen2022wavlm} and Whisper-Large-V3 \citep{radford2023robust}, respectively. 
Next, we retain only those samples whose SpkSim is at least $p$\% of the maximum SpkSim observed among all candidates, ensuring that the selected samples preserve the source speaker’s voice. 
Finally, we pick the sample with the lowest WER among the remaining candidates.

For emotional dialog synthesis, we additionally evaluate the turn alignment score and emotion accuracy. We assign the highest importance to the turn alignment score to reflect its primary role, and we weight emotion accuracy and WER equally.

\section{Experiment}

\subsection{Setups}

\paragraph{\textbf{Training Setup.}}
We use a 3B-parameter model as the default configuration and follow the multi-stage training pipeline described in Section~\ref{sec:train}. In the first pretraining stage, we train on roughly 480k hours of English and Spanish monolingual data with a constant learning rate of $1\times10^{-4}$ for 800k steps on 256 A100 GPUs. In the dubbing SFT stage, we initialize from the pretrained model and fine-tune on about 2k hours of English and Spanish monolingual data plus 50 hours of En$\rightarrow$Es and Es$\rightarrow$En data, using a linearly decaying learning rate starting from $1\times10^{-4}$ for 200k steps on 256 A100 GPUs. In the dialogue SFT stage, we again initialize from the pretrained model and fine-tune on about 28k hours of English full-duplex dialogue data, with a linearly decaying learning rate starting from $1\times10^{-4}$ for 1M steps on 256 A100 GPUs.
For the emotion dialogue SFT, we follow the same training configuration as the dialogue SFT but additionally annotate the data with turn-level valence-arousal-dominance (VAD) labels using an automatic extractor and condition the model on these embeddings.

\paragraph{\textbf{Evaluation Sets.}}
For voice dubbing, we use an internal real-world competitive benchmarking dataset, containing 100 En$\rightarrow$Es and 100 Es$\rightarrow$En samples. For full-duplex dialogue synthesis, we use two test sets: (i) a short-form set with average duration of about 30\,s drawn from a held-out subset of the training data, used for comparison against ground-truth audio, and 
(ii) a long-form set with 1--2\,min text prompts, focusing more on factuality and general helpfulness, used to compare our model with the latest internal system.
For emotional full-duplex dialogue synthesis, we generate 200 multi-turn text dialogues with turn-level emotion labels using an internal text-LLM, spanning four emotion categories (angry, happy, neutral, sad). Then, we condition them on real fully-duplex speech prompts which are held out from training to generate dialogue data for evaluation.

\paragraph{\textbf{Objective Metrics.}}
We evaluate synthetic speech using the following objective metrics. \textbf{WER} measures the content accuracy of the generated speech as the word error rate between the ASR transcript of the generated speech and the target text, using Whisper-Large-V3 \citep{radford2023robust}; lower is better. Note that ASR may still make recognition errors, especially for speech with strong accents, low recording quality, or noisy conditions. \textbf{SpkSim} measures the speaker similarity of the generated speech with respect to the audio prompt, using a WavLM-based speaker verification model~\citep{chen2022wavlm}; higher is better. \textbf{Aes} measures the overall audio quality of the generated speech using the Audiobox-Aesthetics-PQ model~\citep{tjandra2025meta}; higher is better. For full-duplex dialogue synthesis, we report each metric averaged over the two channels.

For emotional full-duplex dialogue synthesis, we additionally report \textbf{turn alignment}, \textbf{emotion accuracy}, and an LLM-based \textbf{dialog naturalness score}. More specifically, we define the turn alignment score to quantify how well the ordering of conversational turns in a generated dialogue matches the intended sequence. After segmenting speech into turns, we compare the synthesized turns with the reference turns using Kendall’s tau correlation coefficient. To evaluate emotion accuracy, we apply Qwen2-Audio\footnote{Qwen2-Audio: \url{https://huggingface.co/Qwen/Qwen2-Audio-7B-Instruct}}~\citep{Qwen2-Audio} as a speech emotion recognizer to classify the emotion of each generated turn after turn segmentation, and compute emotion accuracy against the conditioned emotion labels.
To evaluate the dialog naturalness score, we prompt Gemini~2.5~Pro\footnote{Gemini~2.5~Pro: gemini-2-5-pro-genai-vertex}~\citep{Gemini} with the generated audio to produce a numeric naturalness score on a 1–5 scale, assessing whether the conversation sounds human-like in terms of turn-taking, backchannels, and interjection timing while ignoring conversational content.

\paragraph{\textbf{Subjective Metrics.}}
For voice dubbing, we conduct side-by-side comparisons and collect human ratings on translation accuracy, audio degeneration, prosody similarity, voice similarity, voice naturalness, and shareability. Raters assign a score in the range $[-3, 3]$ for each metric, where a positive score indicates that \our\ is preferred over the comparison system. 
For full-duplex dialogue synthesis, we design a human evaluation protocol based on multi-turn conversations with pairwise comparisons between two variants. Labelers are presented with two conversations side by side, with randomized left/right ordering. The evaluation covers overall human-likeness as well as more fine-grained aspects, including intonation, pacing, expressive intensity, expressive correctness, and use of non-speech vocalizations (NSVs) and fillers. For long-duration samples, we randomly crop each conversation into 1-minute segments to avoid labeler fatigue. We found that using a single randomly sampled segment per conversation is sufficient and therefore do not require annotating all segments from the same sample.

For emotional full-duplex dialogue synthesis, we additionally report MOS scores (scale: 1.0-5.0) for dialog naturalness, emotional interaction smoothness, and emotion alignment. For dialog naturalness, annotators judge whether the generated dialogs resemble human conversations. For emotional interaction smoothness, annotators assess the naturalness of emotional transitions across turns, where higher scores indicate coherent transitions and lower scores indicate abrupt or unjustified changes. For emotion alignment, annotators evaluate how well the expressed emotion in turn-level speech matches the provided emotion label.

\paragraph{\textbf{Inference Setup.}}
For voice dubbing evaluation, we use 32 ODE steps and a reranking size of 32 by default. For full-duplex dialogue synthesis evaluation, we use 32 ODE steps and a reranking size of 4. 
For the SpkSim filtering, we retained the samples whose SpkSim is at least 75\% of the maximum SpkSim score. 
For long-form generation with multi-diffusion, we use a 30\,s chunk size and a 20\,s chunk overlap as the default setup.
For emotional full-duplex dialogue synthesis evaluation, we use 32 ODE steps and a reranking size of 64. 
To control emotion, the user provides a target 
emotion label per turn, which is converted to a VAD vector via a 
predefined mapping, and the model generates emotionally expressive 
dialogue accordingly.

\subsection{Voice Dubbing Evaluation}

\begin{table}[h]
    \centering
    \caption{Human evaluation results for voice dubbing compared to the latest internal dubbing model. Positive scores indicate that \our\ is preferred over the baseline.}
    \label{tab:dub_res_main}
    \begin{tabular}{lcccc}
        \toprule
        \cmidrule(lr){2-3}
        & Es$\rightarrow$En & En$\rightarrow$Es \\
        \midrule
        Translation Accuracy $\uparrow$  &  0.02 &  0.03 \\
        Audio Degeneration $\uparrow$    &  0.16 &  0.06 \\
        Prosody Similarity $\uparrow$    &  0.33 &  0.34 \\
        Voice Similarity $\uparrow$      &  0.29 &  0.36 \\
        Voice Naturalness $\uparrow$     &  0.39 &  0.45 \\
        \midrule
        Shareability $\uparrow$          &  0.40 &  0.38 \\
        \bottomrule
    \end{tabular}
\end{table}

\paragraph{\textbf{Main Results.}}
As summarized in Table~\ref{tab:dub_res_main}, \our\ significantly outperforms the latest internal dubbing model (+0.40 for Es$\rightarrow$En and +0.38 for En$\rightarrow$Es) in shareability by a large margin. 
In particular, \our\ improves over the internal dubbing model on all fine-grained aspects, including translation accuracy, audio degeneration, prosody similarity, voice similarity, and voice naturalness.

\begin{table}[h]
    \centering
    \caption{Ablation on model size for voice dubbing.}
    \label{tab:dub_ab_size}
    \begin{tabular}{lcccc}
        \toprule
        Model size & WER (\%) $\downarrow$ & SpkSim $\uparrow$ & Aes $\uparrow$ \\
        \midrule
        300M  &  44.45 &  0.64 &  6.03 \\
        1B    &  22.97 &  0.72 & 6.26 \\
        3B    &  \textbf{13.98} &  \textbf{0.76} &  \textbf{6.29} \\
        \bottomrule
    \end{tabular}
\end{table}

\paragraph{\textbf{Ablation Study.}}
Table~\ref{tab:dub_ab_size} shows an ablation on model size for the voice dubbing task. 
We observe strong scalability---performance of \our\ consistently improves when increasing the model size from 300M to 1B to 3B parameters. Specifically, compared to the 300M model, the 3B model achieves relative improvements of 69\%, 19\% and 4\% on content accuracy, speaker similarity and audio quality, respectively.

\begin{table}[h]
    \centering
    \caption{Ablation on dubbing SFT.}
    \label{tab:dub_ab_sft}
    \begin{tabular}{lcccc}
        \toprule
          & WER (\%) $\downarrow$ & SpkSim $\uparrow$ & Aes $\uparrow$ \\
        \midrule
        Pre-Training  &  5.93 &  0.69 &  6.72 \\
        mono-lingual SFT    &  3.16 &  \textbf{0.73} &  6.75 \\
        mono$+$cross-lingual SFT    &  \textbf{2.72} &  0.60 &  \textbf{6.82} \\
        \bottomrule
    \end{tabular}
\end{table}
Table~\ref{tab:dub_ab_sft} compares the pretraining-only model with models further fine-tuned using different SFT data configurations. SFT substantially improves all objective metrics, especially WER, indicating much better content accuracy. More specifically, adding cross-lingual data on top of monolingual SFT yields further gains in content accuracy and audio quality, at the cost of a reduction in speaker similarity.

\begin{figure*}[h]
	\centering
	\includegraphics[width=0.6\textwidth]{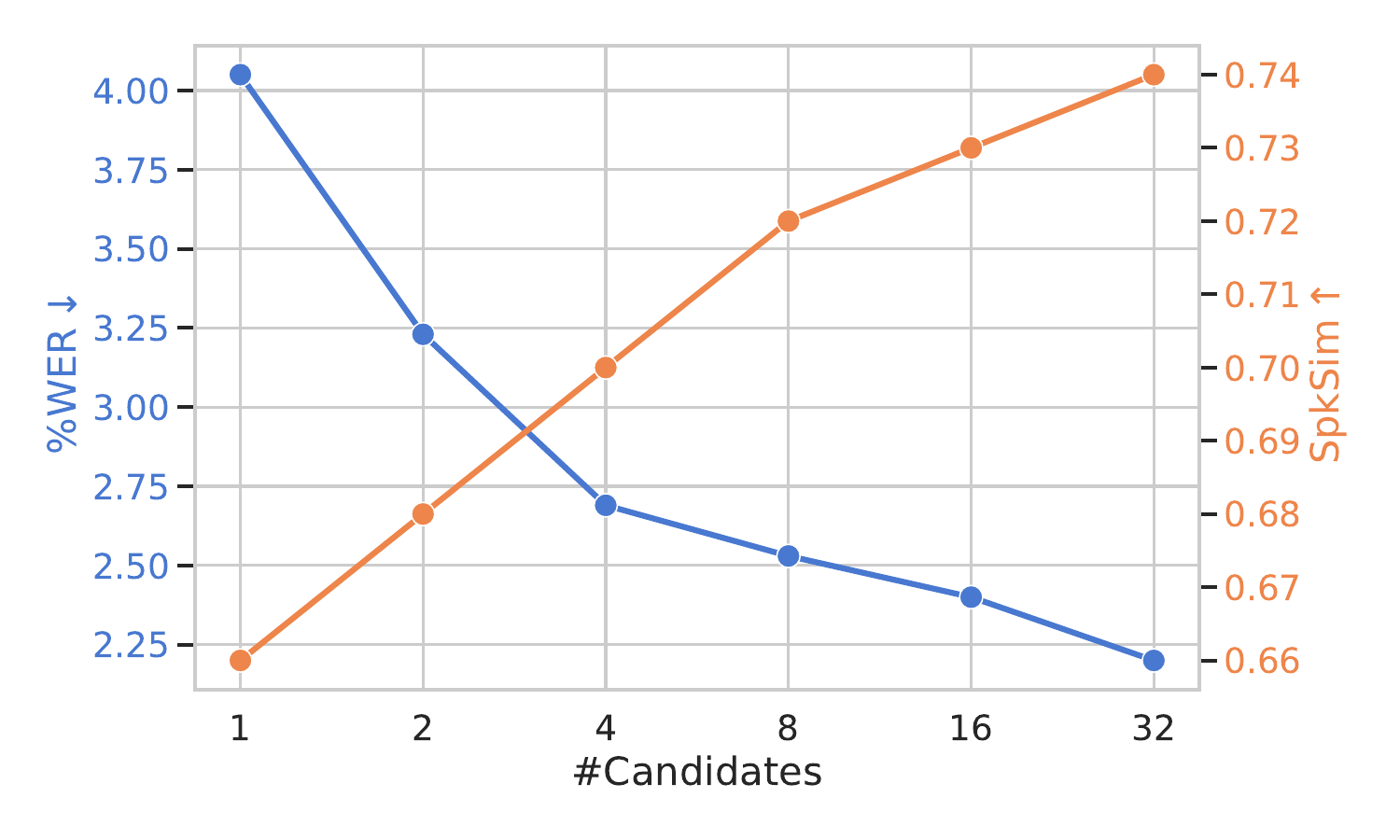}
	\caption{Ablation on multi-stage reranking for voice dubbing.
\label{fig:dub_rerank}
    }
\end{figure*}%

Figure~\ref{fig:dub_rerank} demonstrates that voice dubbing performance is significantly enhanced by the multi-stage reranking strategy, with gains scaling consistently as the number of reranking candidates increases. For instance, using 32 candidates at inference improves the WER from 4.05\% to 2.20\% and the SpkSim from 0.66 to 0.74, compared to a baseline without reranking.

\subsection{Full-Duplex Dialogue Synthesis Evaluation}

\begin{table}[h]
    \centering
    \caption{Human evaluation results for full-duplex dialogue synthesis compared to the latest internal model and ground-truth (GT). MOS scores range from 1 to 5.}
    \label{tab:dia_res_main}
    \begin{tabular}{lcccc}
        \toprule
        & \multicolumn{2}{c}{Short-Form} & \multicolumn{2}{c}{Long-Form}  \\
        \cmidrule(lr){2-3} \cmidrule(lr){4-5}
        & Text-AB & GT & Text-AB & Internal model  \\
        \midrule
        Intonation $\uparrow$            &  4.61 & 4.67 & 3.98 & 3.28 \\
        Pacing $\uparrow$                &  4.52 & 4.62 & 3.73 & 3.39 \\
        Expressive Intensity $\uparrow$  &  4.50 & 4.54 & 3.93 & 3.32 \\
        Expressive Correctness $\uparrow$&  4.62 & 4.70 & 3.96 & 3.62 \\
        NSVs and Fillers $\uparrow$      &  4.76 & 4.81 & 4.12 & 4.57 \\
        \midrule
        Human Likeness $\uparrow$        &  4.53 & 4.62 & 3.86 & 3.00 \\
        \bottomrule
    \end{tabular}
\end{table}

\paragraph{\textbf{Main Results.}}
Table~\ref{tab:dia_res_main} reports human evaluation results for full-duplex dialogue synthesis compared to the latest internal dialogue synthesis model and to ground-truth results. On the short-form evaluation set, \our\ produces dialogues that closely approach real data, with an average gap of only $-0.09$ on overall human-likeness, and similarly strong scores across all fine-grained aspects. On the long-form evaluation set, \our\ significantly outperforms the internal model in terms of overall human-likeness. 
For the more fine-grained dimensions, \our\ also substantially outperforms the internal model on all aspects except NSVs and fillers, likely due to a mismatch between the content domain of the training conversations and that of the evaluation scripts.

\begin{figure*}[h]
	\centering
	\includegraphics[width=1.0\textwidth]{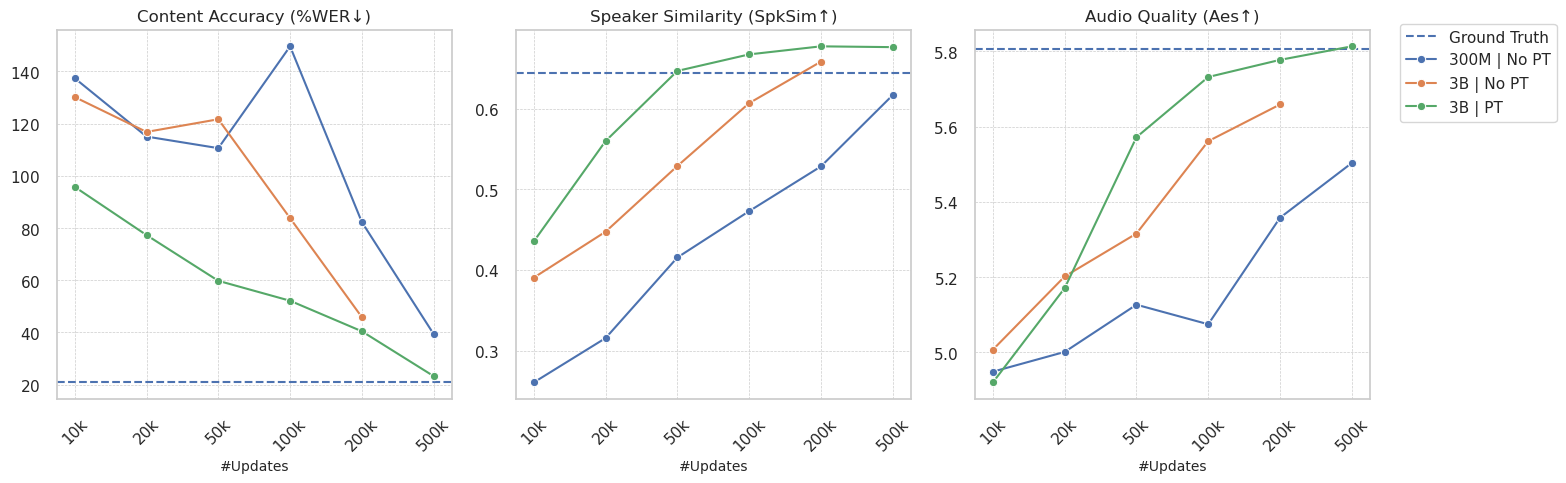}
	\caption{Ablation on model size, training steps, and initialization for full-duplex dialogue synthesis.}
    \label{fig:dia_size}
\end{figure*}%

\paragraph{\textbf{Ablation Study.}}
Figure~\ref{fig:dia_size} presents an ablation study of Text-AB across different model sizes, numbers of training steps, and initialization schemes. We observe that the 3B model significantly outperforms the 300M model on all metrics. Initializing from the pretrained Text-AB-Mono model yields much better performance than training from scratch, across all objective metrics. Moreover, continuing training for more updates consistently improves the results. As a result, a 3B model initialized from the pretrained checkpoint and trained for 500k updates already achieves performance comparable to ground-truth in terms of content accuracy, speaker similarity, and audio quality.

\begin{figure*}[h]
	\centering
	\includegraphics[width=1.0\textwidth]{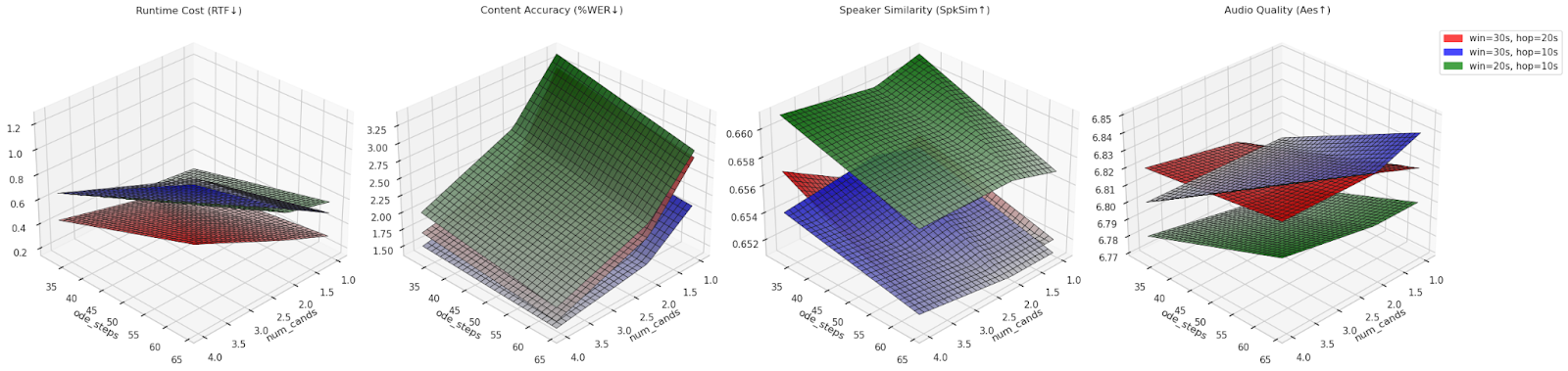}
	\caption{Ablation on multi-diffusion configurations (chunk size, overlap size, number of candidates, and ODE steps) for full-duplex dialogue synthesis.}
    \label{fig:dia_md}
\end{figure*}%

Figure~\ref{fig:dia_md} shows an ablation of multi-diffusion configurations, varying chunk size, overlap size, number of reranking candidates, and ODE steps. Increasing the number of ODE steps and reranking candidates generally improves WER and Aes, but at the cost of a higher real-time factor (RTF). Speaker similarity is comparatively insensitive to these two hyperparameters. Overall, a 30\,s chunk size with 20\,s overlap offers the best trade-off across all four metrics.

\subsection{Emotional Full-Duplex Dialogue Synthesis Evaluation}

\label{sec:emodialogtts_evaluation}

\paragraph{\textbf{Main Results.}}

We compare four models: dialogue SFT, emotional dialogue SFT, MoonCast~\citep{ju2025mooncast}, and CosyVoice2~\citep{CosyVoice2}-based utterance-level concatenation system. 
Dialogue SFT and emotional dialogue SFT are our proposed models, where the latter incorporates explicit turn-level emotion style embeddings to control emotional expression. 
MoonCast addresses a related task---speech dialogue generation from turn-level transcriptions---but differs in two key aspects: it produces single-channel speech containing two speakers and conditions on utterance-level speaker prompts without overlap. 
The CosyVoice2-based baseline independently generates each utterance by prompting CosyVoice2, a zero-shot speech generation model, with emotional reference speech to imitate the target emotion style, then concatenates the resulting utterances to form two-channel dialogues.

We evaluate along three dimensions. For emotion expressiveness, we measure the alignment between the conditioned emotion and the predicted emotion. As shown in Table~\ref{table:emotion_expressiveness}, emotional dialogue SFT consistently outperforms dialogue SFT in both objective and subjective evaluations, demonstrating stronger alignment between intended and generated emotions.

For dialogue naturalness, Table~\ref{table:dialog_naturalness} presents scores obtained from both LLM-based and human evaluations. Using ground-truth text dialogues, emotional dialogue SFT achieves the highest LLM-based naturalness score and the second-highest human evaluation score, indicating that explicit emotion conditioning improves perceived naturalness while maintaining strong human preference. Using synthetic emotional text dialogues, our models achieve the top two scores across both evaluation protocols, and emotional dialogue SFT attains the highest human evaluation score. These results further confirm that incorporating explicit emotion conditioning enhances dialogue naturalness under emotional settings.

For emotional interaction smoothness, we assess the quality of user-agent interaction conveyed through emotional acoustic cues, where human annotators evaluate whether emotional expressions across dialogue turns evolve naturally and coherently. As shown in Table~\ref{table:emotional_interaction}, emotional dialogue SFT achieves the highest score, demonstrating its superior ability to generate natural and emotional interactions.

\begin{table}[h]
\caption{
    Results of emotion expressiveness evaluation. Bold and underline indicate the best and second-best values.
}
\label{table:emotion_expressiveness}
\centering
\scalebox{0.9}{
\begin{tabular}{lcccccc}
\toprule
 & \multicolumn{3}{c}{SER: Qwen2-Audio ($\uparrow$)} & \multicolumn{3}{c}{Emotion Alignment MOS ($\uparrow$)}\\
\cmidrule(lr){2-4}\cmidrule(lr){5-7}
 & angry & happy & sad & angry & happy & sad\\
\midrule
Emotional dialogue SFT & \textbf{0.814} & \textbf{0.428} & \textbf{0.479} & \textbf{3.576}{\tiny $\pm$ {0.252} } & \textbf{3.333}{\tiny $\pm$ {0.380} } & \textbf{3.421}{\tiny $\pm$ {0.288} }\\
Dialogue SFT & \underline{0.650} & \underline{0.396} & \underline{0.455} & \underline{2.397}{\tiny $\pm$ {0.236} } & \underline{1.966}{\tiny $\pm$ {0.295} } & \underline{2.943}{\tiny $\pm$ {0.238} }\\
\bottomrule
\end{tabular}
}
\end{table}

\begin{table}[h]
\caption{
    Results of speech dialog naturalness evaluation. Bold and underline indicate the best and second-best values.
}
\label{table:dialog_naturalness}
\centering
\scalebox{0.85}{
\begin{tabular}{lcccc}
\toprule
 & \multicolumn{2}{c}{Ground-Truth Text} & \multicolumn{2}{c}{Emotional Text} \\
\cmidrule(lr){2-3} \cmidrule(lr){4-5}
 & LLM ($\uparrow$) & MOS ($\uparrow$) & LLM ($\uparrow$) & MOS ($\uparrow$) \\
\midrule
Ground-Truth & 4.844{\tiny $\pm$ {0.033} } & 3.851{\tiny $\pm$ {0.188} } & -- & -- \\
\addlinespace[0.1em]\cdashline{1-5}\addlinespace[0.3em]
Emotional dialogue SFT & \textbf{4.643}{\tiny $\pm$ {0.046} } & \underline{3.906}{\tiny $\pm$ {0.163} } & \underline{3.653}{\tiny $\pm$ {0.073} } & \textbf{3.866}{\tiny $\pm$ {0.088} } \\
Dialogue SFT & \underline{4.468}{\tiny $\pm$ {0.053} } & 3.714{\tiny $\pm$ {0.189} } & \textbf{3.680}{\tiny $\pm$ {0.064} } & \underline{3.090}{\tiny $\pm$ {0.112} } \\
\addlinespace[0.1em]\cdashline{1-5}\addlinespace[0.3em]
CosyVoice2~\citep{CosyVoice2} & 4.285{\tiny $\pm$ {0.059} } & 3.867{\tiny $\pm$ {0.203} } & 2.698{\tiny $\pm$ {0.081} } & 3.009{\tiny $\pm$ {0.181} } \\
MoonCast~\citep{ju2025mooncast} & 4.436{\tiny $\pm$ {0.056} } & \textbf{3.990}{\tiny $\pm$ {0.178} } & 2.960{\tiny $\pm$ {0.088} } & 2.930{\tiny $\pm$ {0.129} } \\
\bottomrule
\end{tabular}
}
\end{table}

\begin{table}[h]
\caption{
    Results of emotional interaction evaluation. Bold and underline indicate the best and second-best values.
}
\label{table:emotional_interaction}
\centering
\scalebox{0.9}{
\begin{tabular}{lcc}
\toprule
  & \multicolumn{1}{c}{Emotional Interaction MOS ($\uparrow$)}\\
\midrule
Emotional dialogue SFT & \textbf{3.675}{\tiny $\pm$ {0.174} } \\
Dialogue SFT & \underline{3.317}{\tiny $\pm$ {0.201} } \\
\bottomrule
\end{tabular}
}
\end{table}

%

\section{Related Work}

\paragraph{\textbf{Zero-Shot TTS.}}

Recent years have witnessed major breakthroughs in zero-shot TTS driven by large-scale model training. VALL-E~\citep{chen2025neural} and a series of follow-up models~\citep{zhang2023speak,kharitonov2023speak,huang2023make,yang2023uniaudio,song2024ella,xin2024rall,lajszczak2024base,chen2024vall} represent speech as discrete codec tokens and formulate TTS as conditional codec language modeling, enabling training on large-scale speech data and performing zero-shot TTS via prompting with strong speaker generalization. Instead of using quantized discrete tokens, more recent work retains this language-modeling formulation but operates on continuous latent features, achieving higher audio quality~\citep{meng2025autoregressive,wang2025felle}. 

In parallel, many approaches leverage non-autoregressive generative models to improve inference speed and stability, including discrete-token prediction methods~\citep{chang2022maskgit,borsos2023soundstorm} and diffusion-based continuous-latent prediction approaches~\citep{li2024styletts,du2024unicats,shen2023naturalspeech,ju2024naturalspeech}. To further improve modeling capacity and audio quality, Audiobox-style models~\citep{le2024voicebox,vyas2023audiobox,eskimez2024e2,anastassiou2024seed,chen2025f5} adopt flow matching~\citep{lipman2022flow} for training, achieving strong performance on zero-shot TTS and speech infilling tasks. Our work follows this Audiobox line of research but extends it in several ways: we move to a high-fidelity latent diffusion space with DAC-VAE features~\citep{polyak2024movie}, remove the need for forced alignment via an alignment-free text interface, and scale model size, data, and training to obtain a unified framework that supports both high-quality voice dubbing and full-duplex dialogue synthesis.

\paragraph{\textbf{Voice Dubbing.}}
Voice dubbing aims to translate speech into a different language while preserving the speaker’s voice, emotion, and expressiveness from a source audio prompt. Practical dubbing systems typically adopt a cascaded pipeline~\citep{yang2020large,federico2020speech,artioli2025generative}: an ASR model transcribes the source speech, a machine translation system produces the target-language transcript, and a TTS model generates the target speech conditioned on speaker characteristics inferred from the source audio. With recent advances in ASR~\citep{li2022recent,zhang2023google,omnilingual2025omnilingual} and machine translation~\citep{wang2022progress,barrault2023seamlessm4t,nllb2024scaling}, the TTS component has become the main remaining bottleneck. Conventional approaches either train a TTS model specifically for each speaker~\citep{yang2020large,federico2020speech} or explicitly extract speaker attributes (e.g., timbre, style, emotion) or other pre-trained features from the source speech and then perform TTS conditioned on these representations~\citep{cong2023learning,cong2024styledubber,artioli2025generative}. Inspired by recent breakthroughs in zero-shot TTS, more recent work instead preserves the source speaker’s characteristics implicitly by conditioning directly on the raw source speech~\citep{sung2025voicecraft}. 
Our work follows this latter line and treats dubbing as a cross-lingual zero-shot TTS problem: we train a large-scale model that conditions on source audio and text, and show that it achieves state-of-the-art dubbing performance.

\paragraph{\textbf{Full-Duplex Dialogue Synthesis.}}
A common approach to dialogue synthesis is to generate each turn as an independent monologue and then concatenate the turns into a conversation~\citep{guo2021conversational,xue2023m,lee2023dailytalk,hu2024fctalker,liu2024generative,xie2025fireredtts}. While this strategy can produce high-quality, natural speech within each turn, it often yields unnatural interactions, weak turn coordination, and limited control over conversational dynamics.
Recent work has therefore shifted toward models that generate entire dialogues in a more integrated manner, explicitly modeling multi-speaker interactions. This includes autoregressive token-prediction approaches~\citep{zhang2024covomix,sesame,notebooklm,darefsky2024parakeet,ju2025mooncast,peng2025vibevoice} and fully non-autoregressive flow-matching-based methods~\citep{zhang2025covomix2}. 

Despite this progress, most of these systems operate in a single-channel setting and do not produce stereo full-duplex dialogue, which is crucial for building high-quality training data for full-duplex speech LLMs~\citep{defossez2024moshi,cui2025recent}.
dGSLM-style models~\citep{nguyen2023generative,mitsui2023towards,lu2025slide} address this by using dual-tower Transformer architectures to capture interleaved speaker information and generate two-channel spoken dialogue autoregressively. ZipVoice-Dialog~\citep{zhu2025zipvoice} further proposes strategies for stereo full-duplex dialogue generation with a non-autoregressive flow-matching-based model. However, existing stereo full-duplex dialogue systems either lack zero-shot voice prompting capabilities or are trained on relatively limited data with modest model scales. In contrast, our work supports zero-shot audio prompting and scales to a 3B-parameter model trained on tens of thousands of hours of two-channel full-duplex dialogue data, targeting high-quality, controllable full-duplex dialogue synthesis.

\paragraph{\textbf{Emotional Speech Synthesis.}}

Prior work on emotional speech generation primarily conditions emotion using categorical labels~\citep{wu2019end} or style embeddings extracted from reference speech~\citep{he2022improve}. 
Subsequent studies introduce continuous scalar variables to control emotion intensity~\citep{zhu2019controlling,zhou2023emotion}, further enabling mixed or compound emotion synthesis through relative attributes~\citep{zhou2022speech,inoue2024hierarchical,inoue2025hierarchical} or diffusion-based interpolation~\citep{tang2023emomix}. 
To improve interpretability and fine-grained control of generated emotion, recent approaches explore structured emotion representations, including explicit prosody modeling~\citep{luo2021emotion,oh2023semi,ren2021fastspeech} and the VAD~\citep{sivaprasad2021emotional,zhou2025emotional,habib2019semi,cho2024emosphere,cho2025emosphere}, which represents emotion along three continuous dimensions: valence, arousal, and dominance~\citep{russell1977evidence}. 
However, all of these methods focus on single-utterance or monologue synthesis and do not address multi-turn dialogue generation. 
Our emotional dialogue SFT extends VAD-based emotion conditioning to two-channel full-duplex dialogue synthesis, enabling turn-level emotion control within natural conversations.

\section{Conclusion}

We introduced \our, an alignment-free TTS framework for voice dubbing, full-duplex dialogue synthesis, and emotional full-duplex dialogue synthesis that unifies single- and two-channel speech generation within a single flow-matching-based DiT architecture. 
\our\ successfully simplified the modeling stack while improving expressivity and scalability by operating in a high-fidelity latent diffusion space with DAC-VAE features, directly conditioning on raw text via cross-attention, and substantially scaling model and data size. 
We first pretrained \our\ on a large-scale multilingual monologue dataset, then fine-tuned on three downstream tasks---dubbing SFT, dialogue SFT, and emotional dialogue SFT.
This enabled a 3B-parameter model to generalize across monolingual, cross-lingual, and two-channel dialogue settings. At inference time, multi-stage reranking strategy enabled high-quality generation, and a multi-diffusion scheme extended one-shot generation to arbitrarily long-form audio with seamless transitions.

Empirically, \our\ sets a new bar for both voice dubbing and full-duplex dialogue synthesis in our production setting. 
It substantially outperformed our internal dubbing model in human evaluations, and produced dialogues that closely match real recordings on short-form benchmarks while markedly improving human-likeness and expressivity over the prior internal dialogue synthesis system on long-form tasks. 
Beyond these results, we believe the alignment-free latent-diffusion design and the multi-diffusion inference strategy provide a general recipe for scalable, controllable speech generation. Future directions include extending \our\ to more languages, modalities, and tasks, further scaling model and data size, and exploring finer-grained controls over style, emotion, and conversational structure.

\clearpage
\newpage
\bibliographystyle{assets/plainnat}
\bibliography{paper}


\end{document}